\documentclass[times,review,10pt]{elsarticle}
\usepackage{amssymb}
\usepackage{amsmath}
\usepackage{amsfonts}
\usepackage{algorithmic}
\usepackage{algorithm}
\usepackage[caption=false,font=normalsize,labelfont=sf,textfont=sf]{subfig}
\usepackage{textcomp}
\usepackage{stfloats}
\usepackage{url}
\usepackage{verbatim}
\usepackage{color}
\usepackage[table]{xcolor}
\usepackage{booktabs}
\usepackage{multirow}
\usepackage{array}
\usepackage{graphicx}
\usepackage{pifont}
\usepackage{bbding}
\usepackage{subfig}
\usepackage{pifont}         % For \ding
\usepackage{fancyhdr}
\usepackage{natbib}
 
\usepackage{cleveref}
\Crefname{figure}{Fig}{Figures}
\journal{Pattern Recognition}
\begin{document}
\begin{frontmatter}
\title{Gesplat: Robust Pose-Free 3D Reconstruction via Geometry-Guided Gaussian Splatting}
\author[1]{Jiahui Lu}
\ead{aujiahui@mail.scut.edu.cn}
\author[1]{Haihong Xiao}
\ead{auhhxiao@mail.scut.edu.cn}
\author[1]{Xueyan Zhao}
\ead{auxyzhao@scut.edu.cn}
%\author[1]{Wenxiong Kang\corref{cor1}}
\author[1]{Wenxiong Kang}
\ead{auwxkang@scut.edu.cn}
%% Author name
%% Author affiliation
%\affiliation[1]{organization={School of Automation Science and Engineering, South China University of Technology},%Department and Organization
            %addressline={381 Wushan Road, Tianhe District}, 
            %city={Guangzhou},
           % postcode={510641}, 
            %state={Guangdong},
            %country={China}}
%\cortext[cor1]{Corresponding author}
%% Abstract
\begin{abstract}
%% Text of abstract
Neural Radiance Fields (NeRF) and 3D Gaussian Splatting (3DGS) have advanced 3D reconstruction and novel view synthesis, but remain heavily dependent on accurate camera poses and dense viewpoint coverage. These requirements limit their applicability in sparse-view settings, where pose estimation becomes unreliable and supervision is insufficient. To overcome these challenges, we introduce Gesplat, a 3DGS-based framework that enables robust novel view synthesis and geometrically consistent reconstruction from unposed sparse images. Unlike prior works that rely on COLMAP for sparse point cloud initialization, we leverage the VGGT foundation model to obtain more reliable initial poses and dense point clouds. Our approach integrates several key innovations: 1) a hybrid Gaussian representation with dual position-shape optimization enhanced by inter-view matching consistency; 2) a graph-guided attribute refinement module to enhance scene details; and 3) flow-based depth regularization that improves depth estimation accuracy for more effective supervision. Comprehensive quantitative and qualitative experiments demonstrate that our approach achieves more robust performance on both forward-facing and large-scale complex datasets compared to other pose-free methods.
\end{abstract}
%%Research highlights
%\begin{highlights}
%\item Hybrid gaussian representation with dual position-shape optimization is leveraged to enhance geometric consistency,
%\item Graph-guided attribute refinement enables scene detail enhancement,
%\item Flow-based depth regularization allows more effective supervision with improved depth estimation accuracy.
%\end{highlights}
%% Keywords
\end{frontmatter}
\begin{figure*}[htbp]
\centering
\includegraphics[width=12cm,height=3.5cm]{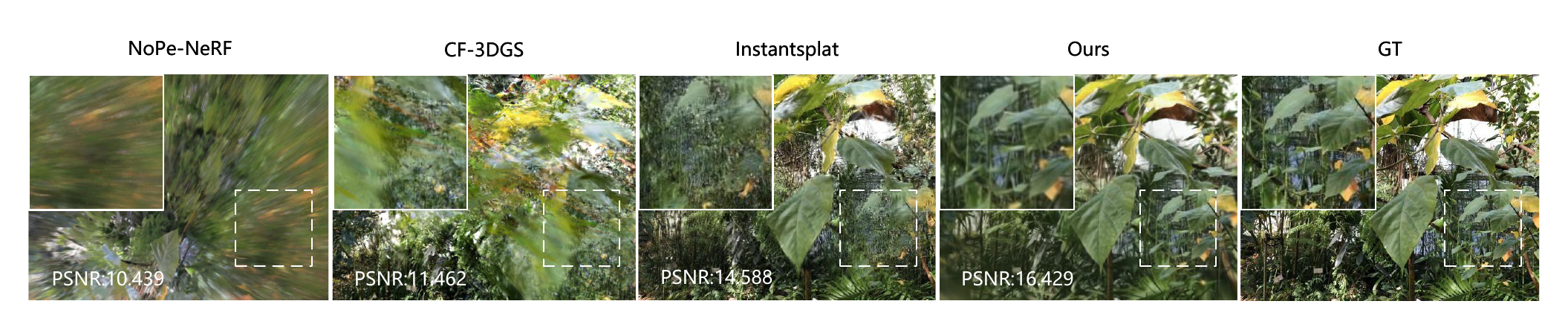}
\caption{Novel View Synthesis Comparisons.We introduce Gesplat, an efficient framework for novel view synthesis and 3D reconstruction from sparse-view unposed inputs.Compared with other pose-free methods, our method achieves higher PSNR and reconstructs more consistent geometry with finer details.
}
\label{fig:leaves}
\end{figure*}
\section{Introduction}
Novel view synthesis and 3D reconstruction are long-standing fundamental goals in computer vision, playing a critical role in applications like autonomous navigation, VR/AR, and robotics. While existing methods can reconstruct scenes from multiple posed images, acquiring dense, well-covered image sets in real-world scenarios is often impractical and costly. The scarcity of views leads to insufficient supervision during training, causing artifacts and flawed reconstructions. Therefore, achieving high-quality 3D reconstruction from only a few sparsely overlapping images remains a critical challenge.

Neural Radiance Fields (NeRF)\cite{mildenhall2021nerf},an implicit neural representation that has shown remarkable success in reconstructing scenes and synthesizing photorealistic novel views. To improve performance under sparse views, NeRF-based methods have attempted in the aspect of pretraining \cite{johari2022geonerf,yu2021pixelnerf}, regularization terms \cite{yang2023freenerf,seo2023mixnerf}, and external priors \cite{wang2023sparsenerf,xie2025tri}, achieving significantly improved results. Nonetheless, these approaches still suffer from high training and rendering costs.

More recently, 3D Gaussian Splatting (3DGS) \cite{kerbl20233d} emerges as an efficient explicit scene representation that employs Gaussian primitives for fast reconstruction and real-time rendering. Subsequent studies \cite{charatan2024pixelsplat, xiong2023sparsegs, cheng2024gaussianpro} further reduce the number of required training views while preserving competitive quality. Nonetheless, these approaches generally assume known and accurate camera poses, which  are rarely satisfied in practical sparse-view settings. In fact, with limited viewpoint overlap, SfM pipelines such as COLMAP \cite{schonberger2016structure} often fail to produce reliable camera parameters. Reconstructing 3D scenes from images without pose information thus remains a challenging yet essential task. Existing pose-free techniques \cite{wang2021nerfmm, bian2023nope, fu2024colmap} eliminate the need for SfM by jointly optimizing scene representation and camera poses end-to-end. However, they typically assume dense video-level coverage and incur high computational costs. InstantSplat \cite{fan2024instantsplat} employs DUSt3R \cite{wang2024dust3r} for coarse geometric initialization, but often trades geometric accuracy for consistency under sparse inputs. To overcome these limitations, we introduce VGGT \cite{wang2025vggt}, a feed-forward model that generates high-fidelity dense point clouds and reliable camera poses. By integrating convolutional inductive biases with self-attention mechanisms, VGGT preserves fine scene details while ensuring global consistency. Its powerful feature matching capability maintains robustness under sparse views, and its strong generalization supports diverse scenes, opening new avenues for 3D reconstruction and pose estimation.

In sparse-view settings, limited input data results in inconsistent scene geometry, and the high interdependence among Gaussian attributes forces the optimization to make a trade-off between optimizing shape and position. Some optimization-based methods \cite{chen2024pgsr,lu2024scaffold} excel in real-time scene reconstruction, but when applied to sparse-view scenarios, they usually introduce invalid geometric priors, leading to incorrect topology and scale ambiguity. Incorporating additional geometric priors is essential to constrain scene structure and improve the accuracy of 3D point positions. Depth information is a commonly used prior due to its connection between 2D images and 3D structure. Some researches \cite{li2024dngaussian,zhu2024fsgs} incorporate monocular depth priors \cite{ranftl2021vision} as regularization, yet such handcrafted constraints often suffer from scale inaccuracy and multi-view inconsistency, which can adversely affect final rendering quality.

In this paper, we propose Gesplat, an effective framework that incorporates appropriate geometric priors to constrain the scene structure while introducing optimization and regularization techniques to refine the scene details. Inspired by \cite{peng2024structure}, we adopt a hybrid Gaussian representation combining ordinary and ray-based Gaussians leveraging matching priors. By binding Gaussians to matching rays, we enhance multi-view consistency and preserve geometric structure. Furthermore, we introduce dual optimization of position and shape to stabilize training and ensure accurate surface convergence. After obtaining a relatively accurate scene representation, we employ a Graph Neural Network (GNN) \cite{xiao2022multi} to further optimize Gaussian attributes. The Gaussians, treated as vertices, are connected based on spatial neighbor relationship, and learnable offsets are applied to update their attributes. To achieve more accurate depth estimation, we utilize an explicit depth computation method within an epipolar geometry framework through optical flow, following \cite{zheng2025nexusgs}. The estimated reliable depth maps are used sequentially for regularization, further improving the rendering quality. During training, we preserve joint training from \cite{fan2024instantsplat} to optimize the Gaussian parameters and camera poses at the same time. Novel view synthesis comparisons with other pose-free methods are shown in Fig.\ref{fig:leaves}, which exhibit the superior performance of our method.

In summary, our method mainly includes the following contributions: 
\begin{itemize}
\item we introduce a hybrid Gaussian representation with dual optimization of position and shape based on matching priors,
\item we design a graph-guided optimization module to refine Gaussian attributes for detailed scene recovery,
\item we apply flow-based depth regularization to improve the quality of rendered images during training,
\item extensive experiments on LLFF and Tanks and Temples datasets show that our method significantly improves scene reconstruction and novel view synthesis from sparse-view pose-free images.
\end{itemize}

\section{Related Work}

\subsection{3D Reconstruction}

As a fundamental challenge in computer vision, 3D reconstruction aims to recover the 3D geometric structure of a scene from a set of input 2D images. Traditional optimization-based methods, such as Structure from Motion (SfM) \cite{schonberger2016structure} and Multi-View Stereo (MVS) \cite{schonberger2016pixelwise,chen2025learning}, rely on rigorous geometric principles to estimate accurate camera parameters and dense point clouds. COLMAP, which integrates SfM and MVS, is one of the most popular and powerful open source tools. However, its high computational cost, sensitivity to errors in SfM-estimated camera poses and sparse point clouds, and dependence on dense viewpoint coverage limit its applicability in textureless regions and areas with low overlap.The advent of deep learning has spurred the development of learning-based reconstruction methods to overcome some limitations of traditional pipelines. As a pioneering work, MVSNet \cite{yao2018mvsnet} constructs a 3D cost volume and uses 3D convolutional neural networks to regress depth maps. Follow-up works such as R-MVSNet \cite{yao2019recurrent} alleviate memory consumption through recurrent networks, and CVP-MVSNet \cite{yang2020cost} introduces a coarse-to-fine pyramid structure to improve efficiency and accuracy. Nonetheless, these approaches typically assume that camera poses are known and accurate, which is usually unreliable in real-world scenarios. More recently, DUSt3R \cite{wang2024dust3r} notably proposes an end-to-end model for feed-forward 3D reconstruction. Unlike traditional methods that require camera calibration or viewpoint poses, DUSt3R leverages a Transformer architecture to directly estimate 3D point clouds from unconstrained image pairs, followed by global alignment to produce a complete scene representation. Its sibling, MASt3R \cite{leroy2024grounding}, augments DUSt3R with a second network head to generate dense local features and introduces a novel matching loss for constraint. To avoid the computational cost of iterative post-optimization required by DUSt3R, VGGT \cite{wang2025vggt} introduces a feed-forward network that directly generates more continuous point clouds and accurate camera poses, capable of processing more than two images simultaneously. VGGT achieves state-of-the-art reconstruction performance compared to DUSt3R and MASt3R. Inspired by this, we adopt VGGT for dense scene initialization.

\subsection{Novel View Synthesis}

Novel view synthesis (NVS) aims to render high-quality images from unseen viewpoints. Neural Radiance Fields (NeRF) \cite{mildenhall2021nerf}, which implicitly represent scenes using multi-layer perceptions (MLPs) and synthesize images via volume rendering, has become one of the most influential NVS methods. Despite achieving high-quality reconstructions, NeRF suffers from high computational costs, leading to slow training and inference. Subsequent research has improved NeRF in various aspects, $e.g.$, quality \cite{barron2022mip,verbin2022ref}, efficiency \cite{garbin2021fastnerf,sun2022direct}, and pose-free \cite{wang2021nerfmm,bian2023nope,cheng2023lu}. More recently, 3D Gaussian Splatting (3DGS) \cite{kerbl20233d} explicitly represents scenes using anisotropic 3D Gaussians \cite{zwicker2001ewa}, enabling rapid reconstruction and real-time rendering while maintaining high visual quality. 3DGS has been extensively explored in many aspects, such as \cite{cheng2024gaussianpro,lu2024scaffold,chen2024hac,lee2024deblurring}, demonstrating superior performance compared with NeRF-based approaches. However, conventional NVS methods heavily rely on dense input views and accurate camera poses derived from SfM, which is often impractical to attain, leading to significant performance degradation under sparse view settings.

\subsection{Sparse-View Novel View Synthesis}

Sparse-view novel view synthesis addresses a practical but more challenging task of synthesizing novel views from a limited set of input images, where overfitting occurs due to insufficient supervision. To mitigate these issues, various strategies have been proposed. Methods like \cite{niemeyer2022regnerf,lee2025sparse} apply regularization techniques to reduce floating artifacts and enhance geometric consistency. \cite{wang2023sparsenerf,deng2022depth} incorporate depth priors from SfM or monocular depth estimation to supervise scene geometry, thereby reducing geometric errors and improving rendering image quality. Generalizable NeRF models like \cite{yu2021pixelnerf,wang2021ibrnet} enable feed-forward inference, allowing fast adaptation to new scenes and improving practicality. Despite these advancements, NeRF-based methods remain computationally intensive due to volume rendering.In contrast to NeRF-based methods, several studies combined with 3DGS attempt to eliminate reconstruction inconsistency. To better capture spatial structure, \cite{li2024dngaussian,zhu2024fsgs} introduce depth information into the explicit 3DGS representation. However, inaccuracies in monocular depth estimation often degrade reconstruction quality in sparse-view scenarios. \cite{xiong2023sparsegs,bao2025loopsparsegs,zhang2024cor} focus on robust scene initialization and removal of incorrectly positioned Gaussians. Some methods \cite{zhang2024cor,liu2024georgs,park2025dropgaussian} introduce various regularization strategies to mitigate geometric degradation and suppress rendering artifacts. Alternative scene representations, such as \cite{peng2024structure,wan2025s2gaussian}, are designed to produce accurate and detailed 3D scenes with enhanced geometry. To better preserve geometric structure completeness and details, we introduce a hybrid Gaussian representation to learn scene structure, with flow-based depth regularization and graph-guided optimization incorporated to refine reconstruction results in weakly textured regions.

\begin{figure*}[htbp]
\centering
\includegraphics[width=12cm,height=5cm]{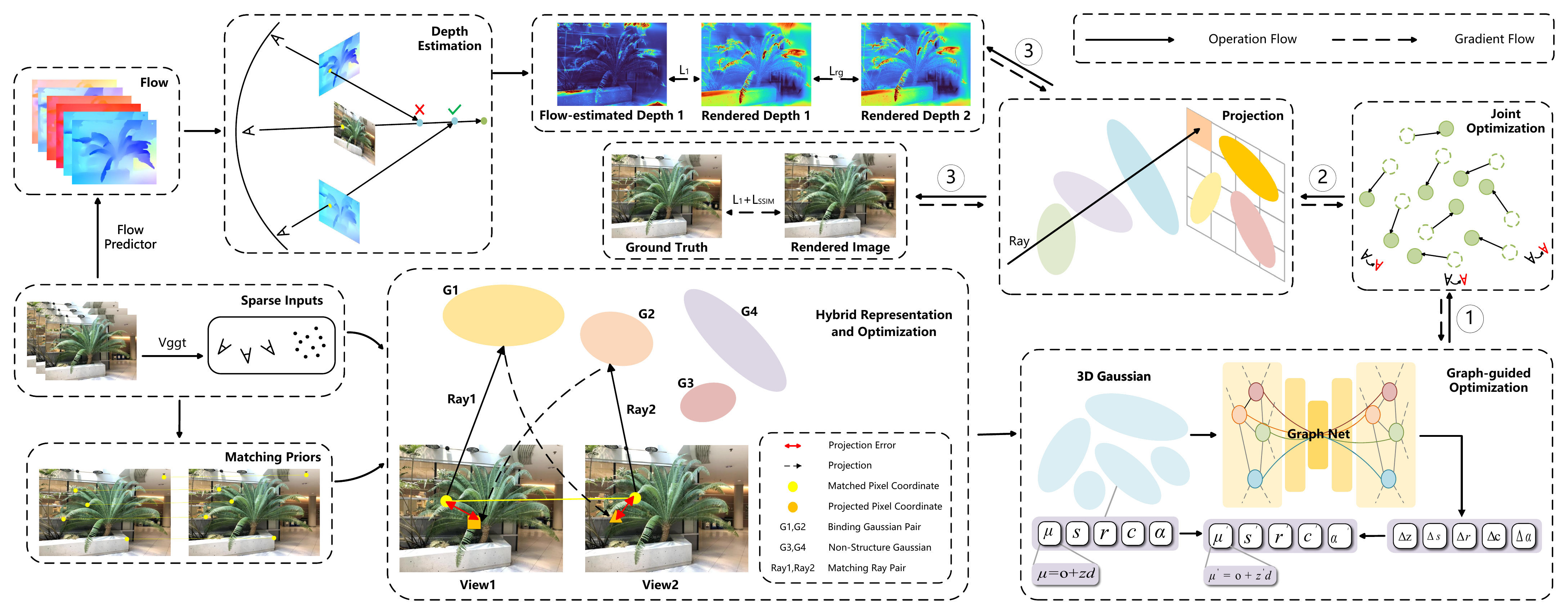}
\caption{Overall framework of Gesplat. Given a few input images, we first generate dense point clouds and camera poses from VGGT, extract matching priors and predict optical flow. Subsequently, we randomly initialize hybrid gaussian representation, using ray-based Gaussians to optimize the gaussian position. Graph-guided optimization and joint optimization are then applied to refine the Gaussians and camera poses. Finally, we employ flow-estimated depth and matching depth as rendering geometry regularization.
}
\label{fig:pipeline}
\end{figure*}

\section{Method}

Given a limited set of RGB images with camera poses and point clouds generated by VGGT, we introduce Gesplat for sparse-view 3D scene reconstruction. Our model consists of four key components: 1) hybrid Gaussian representation, 2) graph-guided optimization, 3) flow-based depth regularization, and 4) joint optimization. Next, we first review the framework of 3DGS, and then provide detailed descriptions of the modules in our method. An overview of Gesplat is illustrated in Fig. \ref{fig:pipeline}.

\subsection{Preliminary}

\subsubsection{Gaussian Splatting}
3DGS\cite{kerbl20233d} explicitly represents a scene through anisotropic 3D Gaussians, achieving high-quality scene reconstruction and real-time rendering. Each Gaussian is parameterized by: a position vector $\mu \in \mathbb{R}^{3}$, a covariance matrix $\Sigma \in \mathbb{R}^{3 \times 3}$, spherical harmonic (SH) coefficients, and an opacity $\alpha \in [0,1)$. The influence for a point $x$ in 3D space is defined as:
\begin{equation}
\label{equ1}
G(x) = e^{-\frac{1}{2}(x - \mu)^{T}\Sigma^{-1}(x - \mu)}.
\end{equation}
To ensure effective computation, the covariance matrix $\Sigma$ is decomposed as $\Sigma = RSS^{T}R^{T}$, where $R \in \mathbb{R}^{3 \times 3}$ is a rotation matrix and $S \in \mathbb{R}^{3 \times 1}$ is a scaling matrix. To enable independent optimization of $R$ and $S$, 3DGS represents them with a rotation factor $r \in \mathbb{R}^{4}$ and a scaling factor $s \in \mathbb{R}^{3}$, ensuring $r$ is normalized to a valid unit quaternion. With a view transformation matrix $W$ and the Jacobian of the affine approximation of the projective transformation $J$, 3D Gaussians are projected to the 2D image plane for rendering:
\begin{equation}
\label{equ2}
\Sigma' = JW\Sigma W^{T}J^{T}.
\end{equation}
Point-based rendering blends $N$ ordered Gaussians overlapping a pixel and computes the color $C$ of the pixel via alpha blending:
\begin{equation}
\label{equ3}
C = \sum_{i \in \mathcal{N}} c_{i}\alpha_{i}'\prod_{j = 1}^{i - 1}(1 - \alpha_{j}'),
\end{equation}
where $c_{i}$ is the color of the $i$-th Gaussian, and $\alpha_{i}'$ is the multiplication of opacity $\alpha_{i}$ and $\Sigma'$. Depth $D$ is computed similarly:
\begin{equation}
\label{equ4}
D = \sum_{i \in \mathcal{N}} d_{i}\alpha_{i}'\prod_{j = 1}^{i - 1}(1 - \alpha_{j}'),
\end{equation}
where $d_{i}$ is the depth of the $i$-th Gaussian.
After rendering, the photometric loss between the rendered image and the ground truth image is computed to optimized the model. All learnable parameters are optimized via stochastic gradient descent using a combination of L1 and D-SSIM \cite{wang2004image} loss:
\begin{equation}
\label{equ5}
L_{photo} = (1 - \lambda)L_{1} + \lambda L_{D-SSIM}.
\end{equation}

\subsubsection{Dense Initialization}

Although 3DGS is efficient in real-time rendering, its initialization relies on a sparse point cloud and accurate camera parameters from Structure-from-Motion (SFM) \cite{schonberger2016structure}. This usually leads to inadequate information in sparse-view scenarios, which causes overfitting on training views and overly smooth textures. Thanks to advances in deep learning, recent learning-based 3D reconstruction methods bypass the need for explicit keypoint extraction and matching of traditional methods in an end-to-end way. Among these, DUSt3R \cite{wang2024dust3r} estimates 3D point clouds directly from 2D image pairs without precise camera calibration, and employs a global alignment strategy for multi-view 3D reconstruction. However, its iterative post-optimization steps result in high computation cost. To address this, we utilize VGGT \cite{wang2025vggt}, a feedforward neural network that produces high-fidelity dense point clouds and accurately estimates camera poses even from sparse and low-overlap datasets. This ensures geometric continuity in the reconstructed scene and provides a more robust initialization for 3DGS.

 \subsection{Hybrid Gaussian Representation and Optimization}
 
 In the sparse-view scenario, 3DGS model struggles to learn the complete scene structure with inconsistent information from training views, leading to poor reconstruction quality in novel views. To better learn the scene geometry, we extract matching priors from a pre-trained matching model \cite{shen2024gim} and introduce a hybrid Gaussian representation. Ray-based Gaussians are bound to matching ray pairs across views and optimized along the ray direction, while ordinary non-structured Gaussians represent background regions visible in individual views.
 
\subsubsection{Matching Priors}

Matching priors refer to ray correspondence and ray position.
1) Ray Correspondence: A pair of matching rays $\{r_i, r_j\}$ from view $i$ and view $j$ corresponds to the same 3D point. Given image $I_i$ and $I_j$, matching ray pair $\{r_i, r_j\}$, pixel points $\{p_i, p_j\}$, intrinsic parameters $\{K_i, K_j\}$, and extrinsic parameters $\{[R_i, t_i], [R_j, t_j]\}$, we assume that each ray intersects the surface at point $X_i$ and $X_j$, satisfying $X_i = X_j$. Thus we can get the projection from view $i$ to view $j$ as:
\begin{equation}
\label{equ6}
{p_{j} = p}_{i \rightarrow j}(X_{i}) = \pi(K_{j}R_{j}^{T}(X_{i} - t_{j})),
\end{equation}
where $\pi([x,y,z]^{T}) = [x/z,y/z]^{T}$. The projection from view $j$ to view $i$ is defined similarly.
2) Ray Position: We emphasize that ray position refers to the positions visible in two or more views. In the sparse-view scenario, non-overlapping regions often cause overfitting. Therefore, regions visible across multiple views are very crucial.

\subsubsection{Ray-based Gaussian}

Given an image pair $I_i$ and $I_j$ with $N$ matching ray pairs $\{r_i^k, r_j^k\}_{k=1}^N$, there exist $N$ corresponding pairs of ray-based Gaussians $\{G_i^k, G_j^k\}_{k=1}^N$. Each primitive has similar attributes with ordinary Gaussian but differs in position representation. The new representation of position $\mu'$ is as follows:
\begin{equation}
\label{equ7}
\mu' = o + z d,
\end{equation}
where $o$ is the camera center, $d$ is the ray direction, and $z$ is a randomly initialized learnable distance factor.
\subsubsection{Position Optimization}
Leveraging the binding strategy between ray-based Gaussians and their matching rays, we optimize Gaussian positions using the projection relation. For a matching ray pair $\{r_i, r_j\}$ with corresponding Gaussians $\{G_i, G_j\}$ at positions $\mu_{i}' = o_{i} + z_{i}d_{i}$ and $\mu_{j}' = o_{j} + z_{j}d_{j}$, the projection error is defined as:
\begin{equation}
\label{equ8}
\left\{
\begin{aligned}
 & L_{gp}^{i \rightarrow j} = \parallel p_{j} - p_{i \rightarrow j}(\mu_{i}') \parallel \\
 & L_{gp}^{j \rightarrow i} = \parallel p_{i} - p_{j \rightarrow i}(\mu_{j}') \parallel .
\end{aligned}
\right.
\end{equation}
The total Gaussian position loss $L_{gp}$ is the average of all projection errors.

\subsubsection{Rendering Geometry Optimization}

In addition to position, other inaccurate attributes can also affect rendering quality. To address this, we optimize the rendering geometry. Through Eq. (4), we obtain the rendering depth maps $D_i$ and $D_j$. Then we use the depth values $\{D_i(p_i), D_j(p_j)\}$ in matching pixels to yield the corresponding 3D points:
\begin{equation}
\label{equ9}
P_{i} = R_{i}(D_{i}(p_{i})K_{i}^{-1}\widetilde{p}_{i})) + t_{i},
\end{equation}
where $\widetilde{p}_{i}$ is the 2D homogeneous coordinate of pixel $p_i$ ($P_j$ is similar). With the projection relation, the rendering depth projection error is given by:
\begin{equation}
\label{equ10}
\left\{
\begin{aligned}
 & L_{rg}^{i \rightarrow j} = \parallel p_{j} - p_{i \rightarrow j}(P_{i}) \parallel \\
 & L_{rg}^{j \rightarrow i} = \parallel p_{i} - p_{j \rightarrow i}(P_{j}) \parallel .
\end{aligned}
\right. 
\end{equation}
The total rendering geometry loss $L_{rg}$ is the average of all these errors.

\subsection{Graph-guided Optimization}

Considering the view inconsistency under a sparse-view setting, we introduce a Graph Neural Network (GNN) to better learn the spatial geometry structure, improving the scene reconstruction quality. We utilize a graph $\mathcal{G = (V, E)}$ to capture complex geometric features for Gaussians, where vertices $\mathcal{V} = \{v_i\}_{i=1}^M = \{z_i, s_i, r_i, c_i, \alpha_i\}_{i=1}^M$ represent the Gaussian attributes, and edges $\mathcal{E}$ describe the spatial adjacency based on K nearest neighbors (KNN). The K nearest neighbors of point $\mu_i$ are calculated as:
\begin{equation}
\label{equ11}
KNN(\mu_i) = \{\mu_j \mid j \in \text{argmin}_{j} \, d(\mu_i, \mu_j),j = 1,\cdots,K\},
\end{equation}
where the Euclidean distance $d(\mu_i, \mu_j) = \parallel \mu_i - \mu_j \parallel_2$. And the edges are defined as:
\begin{equation}
\label{equ12}
\mathcal{E = \{(}i, j) \mid \mu_j \in KNN(\mu_i) \land d(\mu_i, \mu_j) < r\},
\end{equation}
where $r$ is a radius threshold.
Upon generating the graph network, we formulate its output as follows:
\begin{equation}
\label{equ13}
\delta = \mathcal{G}(\mathcal{V, E}) = \{\Delta z_i, \Delta s_i, \Delta r_i, \Delta c_i, \Delta \alpha_i\}_{i=1}^M.
\end{equation}
Then we introduce the output as the refinement of Gaussian attributes:
\begin{equation}
\label{equ14}
\begin{aligned}
z_{i}' &= z_{i} + \lambda_{z} \cdot \Delta z_{i}, \\
s_{i}' &= s_{i} + \lambda_{s} \cdot \Delta s_{i}, \\
r_{i}' &= s_{i} + \lambda_{r} \cdot \Delta r_{i}, \\
c_{i}' &= c_{i} + \lambda_{c} \cdot \Delta c_{i}, \\
\alpha_{i}' &= \alpha_{i} + \lambda_{\alpha} \cdot \Delta \alpha_{i},
\end{aligned}
\end{equation}
and the refined Gaussian position is given by:
\begin{equation}
\label{equ15}
\mu_{i}'' = o_{i} + z_{i}' d_{i}.
\end{equation}
We set the regularization parameters $\lambda_{z}$, $\lambda_{s}$, $\lambda_{r}$, $\lambda_{c}$, and $\lambda_{\alpha}$ to 0.1, 0.1, 0.05, 0.01, and 1, respectively.

\subsection{Flow-based Depth Regularization}

In addition to attaining high reconstruction quality using matching priors and rendering depth, we also introduce the depth estimated through optical flow to calibrate the geometric structure. Since monocular depth estimation often fails to deliver accurate depth information, we incorporate epipolar priors as a constraint to explicitly compute the depth maps.

\subsubsection{Epipolar Line}

For a point $p_i$ in view $i$, the corresponding epipolar line $l_{p_i}$ in view $j$ is defined as:
\begin{equation}
\label{equ16}
\begin{aligned}
a x &+ b y + c = 0, \\
(a, b, c)^{\top} &= F^{i \rightarrow j} (x_i, y_i, 1)^{\top}, 
\end{aligned}
\end{equation}
where $F^{i \rightarrow j}$ is the fundamental matrix from view $i$ to view $j$, and $(x_i, y_i)$ is the coordinate of the point $p_i$.

\subsubsection{Flow-estimated Depth}

Given a pre-trained optical flow predictor $f(\bullet)$ \cite{shi2023flowformer++}, we compute the optical flow between views $i$ and $j$ as:
\begin{equation}
\label{equ17}
M_{flow}^{i \rightarrow j} = f(I_i, I_j),
\end{equation}
where $I_i$ and $I_j$ is a matching image pair. Given a point $p_i$ in image $I_i$, the predicted corresponding point $\widehat{p}_{j}$ in image $I_j$ is estimated as:
\begin{equation}
\label{equ18}
\widehat{p}_{j} = p_{i} + M_{flow}^{i \rightarrow j}(p_{i}).
\end{equation}
Due to estimation errors, $\widehat{p}_{j}$ may not accurately lies on the epipolar line $l_{p_i}$. Therefore, we select the perpendicular foot $\overline{p}_{j}$ as the approximate estimation of $\widehat{p}_{j}$ :
\begin{equation}
\label{equ19}
\overline{p}_{j} = \left( \frac{b^{2} \widehat{x}_{j} - a b \widehat{y}_{j} - a c}{a^{2} + b^{2}}, \frac{a^{2} \widehat{y}_{j} - a b \widehat{x}_{j} - b c}{a^{2} + b^{2}} \right),
\end{equation}
where $(\widehat{x}_{j}, \widehat{y}_{j})$ is the coordinate of point $\widehat{p}_{j}$, and $\{a, b, c\}$ are the parameters of the epipolar line $l_{p_i}$. Then the depth value of the point $p_i$ is calculated as:
\begin{equation}
\label{equ20}
\begin{aligned}
D^{i\to j}(p_i,\bar{p}_j)  &=\frac{|H\times-(R_iR_j^{-1}T_j-T_i)|}{|(K_i^{-1}(x_i,y_i,1)^{\mathsf{T}})\times H|}, \\
\mathrm{where~}H =&(R_jR_i^{-1})^{-1}K_j^{-1}(\bar{x}_j,\bar{y}_j,1)^{\mathsf{T}}.
\end{aligned}
\end{equation}
Here, $\times$ denotes the cross product, and $\{K_i, K_j\}$ and $\{[R_i, t_i], [R_j, t_j]\}$ are the camera intrinsics and extrinsics for view $i$ and view $j$, respectively.

\subsubsection{Cross-view Depth Blending}

Given image set $\{I_i \mid i=1,2,\cdots,N\}$, for a point $p_i$ in $I_i$, we compute the predicted matching points $\{\overline{p}_j \mid j \ne i\}$ and the corresponding depths in other $N-1$ views. Although $\overline{p}_j$ lies on the epipolar line, the distance error that $\overline{p}_j$ deviates from the true correspondence $\widehat{p}_j$ still persists. To handle the variance of depth estimation across views, we introduce a cross-view depth blending strategy.
Depth is represented as the distance information from a pixel point to the camera. Given $O_i$ as the camera center of view $i$, we define a reference distance $dis_{ref}$ as the distance from the 3D point $\overline{P}_j$ to $O_i$. And the projection distance $dis_{pro}$ in view $j$ is the distance from the projected point $\overline{p}_j$ to the epipole $o_{ij}$. The distance between $\overline{p}_j$ and the true matching point $p_j$ is denoted as $\Delta dis_{pro}$, corresponding to a 3D distance $\Delta dis_{ref}$. A smaller $\Delta dis_{ref}$ indicates a more accurate depth estimation. Through the fundamental theorem of calculus, the transformation between $\Delta dis_{ref}$ and $\Delta dis_{pro}$ is computed as:
\begin{equation}
\label{equ21}
\begin{aligned}
\Delta dis_{ref} &= \int_{dis_{pro}}^{dis_{pro} + \Delta dis_{pro}} dis_{ref}'(dis_{pro}) \, d dis_{pro} \\
&\approx dis_{ref}'(dis_{pro}) \Delta dis_{pro},
\end{aligned}
\end{equation}
where $dis_{ref}'(dis_{pro})$ is the gradient of $dis_{ref}(dis_{pro})$, indicating the sensitivity of depth to the changes of $dis_{pro}$. A smaller $dis_{ref}'(dis_{pro})$ implies higher confidence in the estimated depth. Using geometric relations, the formulation of $dis_{ref}'(dis_{pro})$ is expressed as:
\begin{equation}
\label{equ22}
dis_{ref}'(dis_{pro}) = \frac{t \sin \beta \sin^2(\alpha + \theta)}{m \sin \theta \sin^2(\alpha + \beta)},
\end{equation}
where $t = O_i O_j$, $m = O_j o_{ij}$, $\alpha = \angle O_i O_j \overline{P}_j$, $\theta = \angle O_j o_{ij} \overline{p}_j$, $\beta = \angle O_j O_i \overline{P}_j$.Since we seek the view $j$ with the smallest $\Delta dis_{ref}$ from other $N-1$ views, assuming that $\Delta dis_{pro}$ is constant across views for the same optical flow predictor, we only need to compare $dis_{ref}'(dis_{pro})$. Finally, the blended depth $D_{i}(p_{i})$ for point $p_i$ is given by:
\begin{equation}
\label{equ23}
D_{i}(p_{i}) = D^{i \rightarrow k}(p_{i}), \quad k = \arg\min_{j} dis_{ref}'(dis_{pro}^{i \rightarrow j, p_i}).
\end{equation}

\subsubsection{Depth Regularization}

Given the rendered depth map $D$ and the flow-estimated depth map $\widehat{D}$, the depth loss is computed using loss function L1:
\begin{equation}
\label{equ24}
L_{depth} = L_{1}(\widehat{D}, D).
\end{equation}
\subsection{Joint Optimization}
To jointly optimize noisy point clouds and inaccurate camera poses, we introduce joint optimization, which updates both Gaussians and camera parameters via image loss. Given a Gaussian set $G$ and camera parameters $T$, we minimize the difference between the ground truth image and the rendered image using gradient descent:
\begin{equation}
\label{equ25}
G^{*}, T^{*} = \arg\min_{G, T} \sum_{v \in N} \sum_{i=1}^{HW} \parallel \widetilde{C}_{v}^{i}(G, T) - C_{v}^{i} \parallel.
\end{equation}
During testing, since predicted camera poses for novel views may still be noisy, we use the trained Gaussian model to refine test camera poses via Eq.\ref{equ25} for 500 iterations per-image for more accurate evaluation.

\subsection{Training Objectives}

\subsubsection{Loss Function}

The total training loss is formulated as :
\begin{equation}
\label{equ26}
L = L_{photo} + \lambda_{gp}L_{gp} + \lambda_{rg}L_{rg} + \lambda_{depth}L_{depth},
\end{equation}
where $L_{photo}$ is the photometric loss, $L_{gp}$ is the Gaussian position loss, $L_{rg}$ is the rendering geometry loss, and $L_{depth}$ is the depth loss.

\subsubsection{Training Details}

We set $\lambda_{gp} = 1.0$, $\lambda_{depth} = 0$, $\lambda_{rg} = 0$ for the first 2000 iterations, and then adjust to $\lambda_{depth} = 0.1$, $\lambda_{rg} = 0.3$ to avoid sub-optimization during early stage of training . Our model is built upon InstantSplat \cite{fan2024instantsplat} and trained for 5000 iterations. The learning rate for the learnable depth $z$ starts at 0.1 and decays to $1.6 \times 10^{-6}$. To mitigate initial errors and computational cost from the graph network, we only use it for Gaussian optimization in the final 200 iterations.

\section{Experiments}

In this section, we conduct experiments to demonstrate the effectiveness of our approach. We first describe the datasets, evaluation metrics, and baselines used for comparison. Next, we compare our method with other pose-free methods on LLFF \cite{mildenhall2019local} and Tanks and Temples \cite{knapitsch2017tanks} datasets, followed by an analysis of the results. Finally, we perform a series of ablation studies to validate the efficacy of key modules in our method.

\subsection{Experimental Settings}

\subsubsection{Datasets}

Under sparse-view settings, we evaluate Gesplat using LLFF and Tanks and Temples datasets. LLFF provides a collection of real-world scenes with forward-facing camera trajectories, while Tanks and Temples is a large-scale dataset containing both indoor and outdoor scenes with complex environments. Both datasets consist of eight scenes each. Following InstantSplat \cite{fan2024instantsplat}, we randomly sample 24 images from each scene, with 12 images held out as test views (including the first and last images). For the remaining 12 images, we select 3 to 9 images for training under sparse-view conditions.

\subsubsection{Baselines $\&$ Metrics}

We compare Gesplat with other pose-free approaches, including Nope-NeRF \cite{bian2023nope}, CF-3DGS \cite{fu2024colmap}, and InstantSplat \cite{fan2024instantsplat}. Among them, Nope-NeRF and CF-3DGS leverage monocular depth estimation and ground-truth camera intrinsics, while InstantSplat relies on DUSt3R \cite{wang2024dust3r} to generate dense point clouds. All of the baselines are trained using official code with default settings. We evaluate the performance on novel view synthesis task using common metrics: Peak Signal-to-Noise Ratio (PSNR), Structural Similarity Index (SSIM) \cite{wang2004image}, and Learned Perceptual Image Patch Similarity (LPIPS) \cite{zhang2018unreasonable}.

\subsubsection{Implementation Details}

Our implementation is built on the PyTorch framework, and all experiments are conducted on one NVIDIA RTX 3090 GPU. For Nope-NeRF and CF-3DGS, we selete input images at the same resolution as original datasets. Using DUSt3R, InstantSplat predicts multi-view stereo depth maps at a resolution of 512. Our method trains with a resolution of 518 × 392 using VGGT \cite{wang2025vggt}.
\begin{table*}[htbp]
	\centering
	\caption{Quantitative comparison on LLFF ~(3,6,9 views). The best, second best, and third best entries are marked in red, orange, and yellow, respectively. }
	\setlength{\tabcolsep}{2mm}
	\footnotesize
    \resizebox{0.90\textwidth}{!}{
	\begin{tabular}{l|ccc|ccc|ccc}
		\toprule[1pt]   
		\multirow{2}{*}{\textbf{Method}} & \multicolumn{3}{c|}{(3 views)} & \multicolumn{3}{c|}{(6 views)} & \multicolumn{3}{c}{(9 views)}  \\
		%\cmidrule(lr){2-} 
		& PSNR$\uparrow$ & SSIM$\uparrow$ & LPIPS$\downarrow$  & PSNR$\uparrow$ & SSIM$\uparrow$ & LPIPS$\downarrow$ 
        & PSNR$\uparrow$ & SSIM$\uparrow$ & LPIPS$\downarrow$  \\
        \midrule
		NoPe-NeRF\cite{bian2023nope}
        &\cellcolor{yellow! 30}12.029&\cellcolor{yellow! 30}0.326&0.696&12.084&0.332&0.692&11.883&0.330&0.691 \\
		CF-3DGS\cite{fu2024colmap}
        &7.891&0.193&\cellcolor{yellow!30}0.637&\cellcolor{yellow! 30}14.380&\cellcolor{yellow! 30}0.398&\cellcolor{yellow! 30}0.488&\cellcolor{yellow! 30}14.849&\cellcolor{yellow! 30}0.411&\cellcolor{yellow! 30}0.496\\
        InstantSplat\cite{fan2024instantsplat}&\cellcolor{orange!30}16.726&\cellcolor{orange!30}0.465&\cellcolor{red!30}0.355&\cellcolor{orange!30}19.315&\cellcolor{orange!30}0.582&\cellcolor{orange!30}0.280&\cellcolor{orange!30}20.616&\cellcolor{orange!30}0.622&\cellcolor{orange!30}0.255\\
		\textbf{Ours} &\cellcolor{red!30}18.869&\cellcolor{red!30}0.570&\cellcolor{red!30}0.355&\cellcolor{red!30}21.994&\cellcolor{red!30}0.692&\cellcolor{red!30}0.231&\cellcolor{red!30}23.299&\cellcolor{red!30}0.738&\cellcolor{red!30}0.193 \\
		\bottomrule[1pt]
    \end{tabular}}
	\label{tab:llff_comparison}
	\vspace{-7pt}
\end{table*}

\begin{figure*}[htbp]
\centering 
\includegraphics[width=12cm,height=8cm]{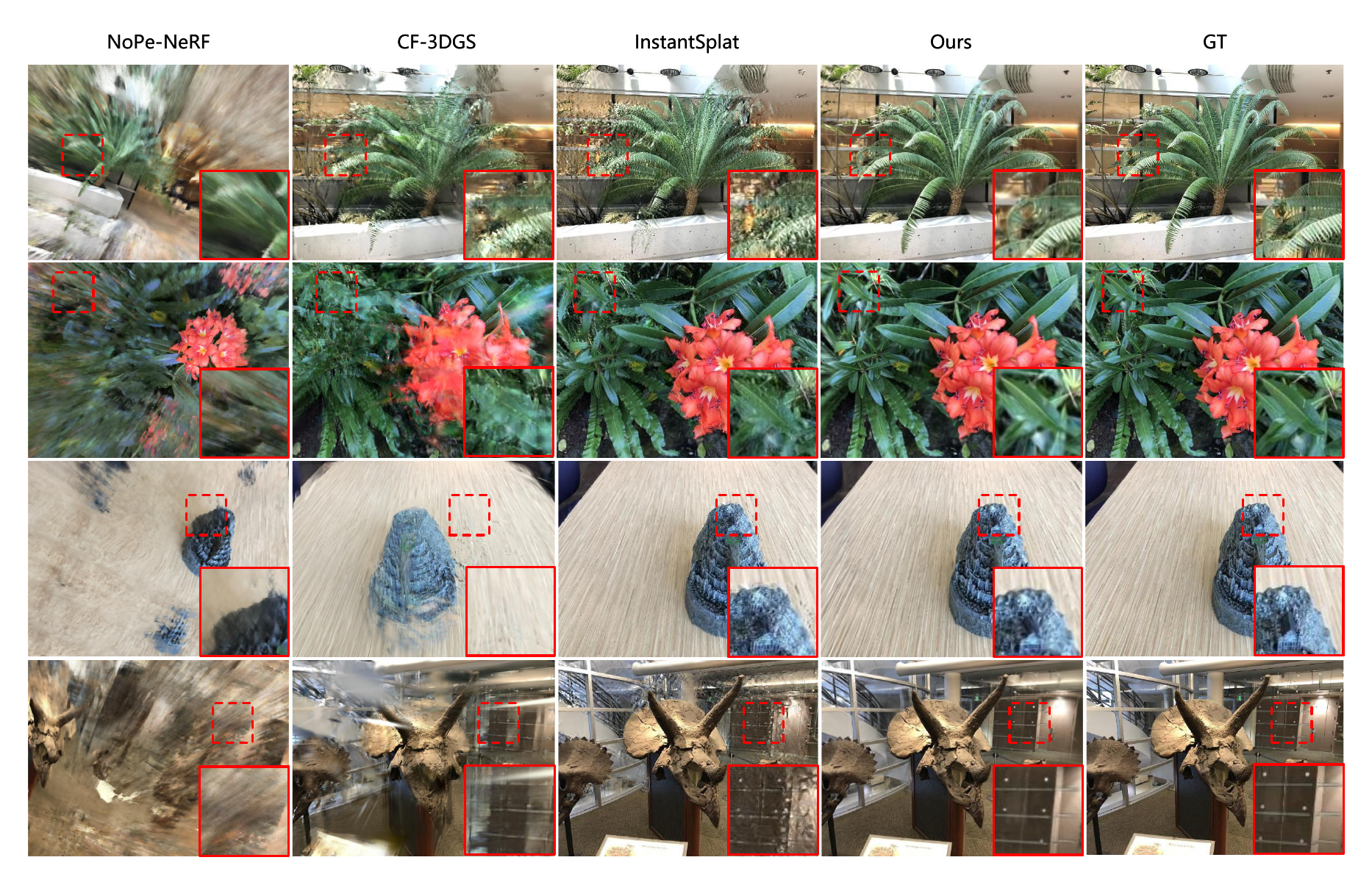}
\caption{ Qualitative comparison between Gesplat and various pose-free methods on LLFF datasets(6 views).The reconstruction of our method is more accurate and exhibits finer details compared with other competitors.}
\label{fig:llff_6}
\end{figure*}

\subsection{Experimental Results and Analysis}

\subsubsection{Results on LLFF}

Quantitative comparisons are summarized in Tab. \ref{tab:llff_comparison}.  Gesplat achieves the best performance across PSNR, SSIM, and LPIPS compared to other pose-free methods. Nope-NeRF utilizes Multilayer Perceptions (MLPs) to prioritize learning the global scene structure, which effectively suppresses floating artifacts but may produce images suffering from over-blur and require a long training time. CF-3DGS builds upon 3DGS with local and global optimization, enabling more accurate geometric reconstruction and real-time rendering, yet its complex optimization process may lead to loss of fine details. As a feed-forward model, InstantSplat uses DUSt3R to estimate dense point clouds and camera poses, mitigating the information scarcity of 3DGS in sparse views and improving geometric consistency. However, it may occasionally generate inexistent details, reducing the reconstruction quality. By leveraging VGGT for dense 3D geometry recovery, our method uses a hybrid Gaussian representation to learn scene geometry and refines Gaussian attributes and camera parameters via a graph network and joint optimization. Furthermore, we incorporate depth cues from optical flow estimation to improve rendering quality. Qualitative results under 6 training views are shown in Fig. \ref{fig:llff_6}, where Gesplat not only reconstructs complete geometry but also recovers more accurate and high-frequency details.

\subsubsection{Results on Tanks and Temples}

\begin{figure*}[htbp]
\centering 
\includegraphics[width=12cm,height=11cm]{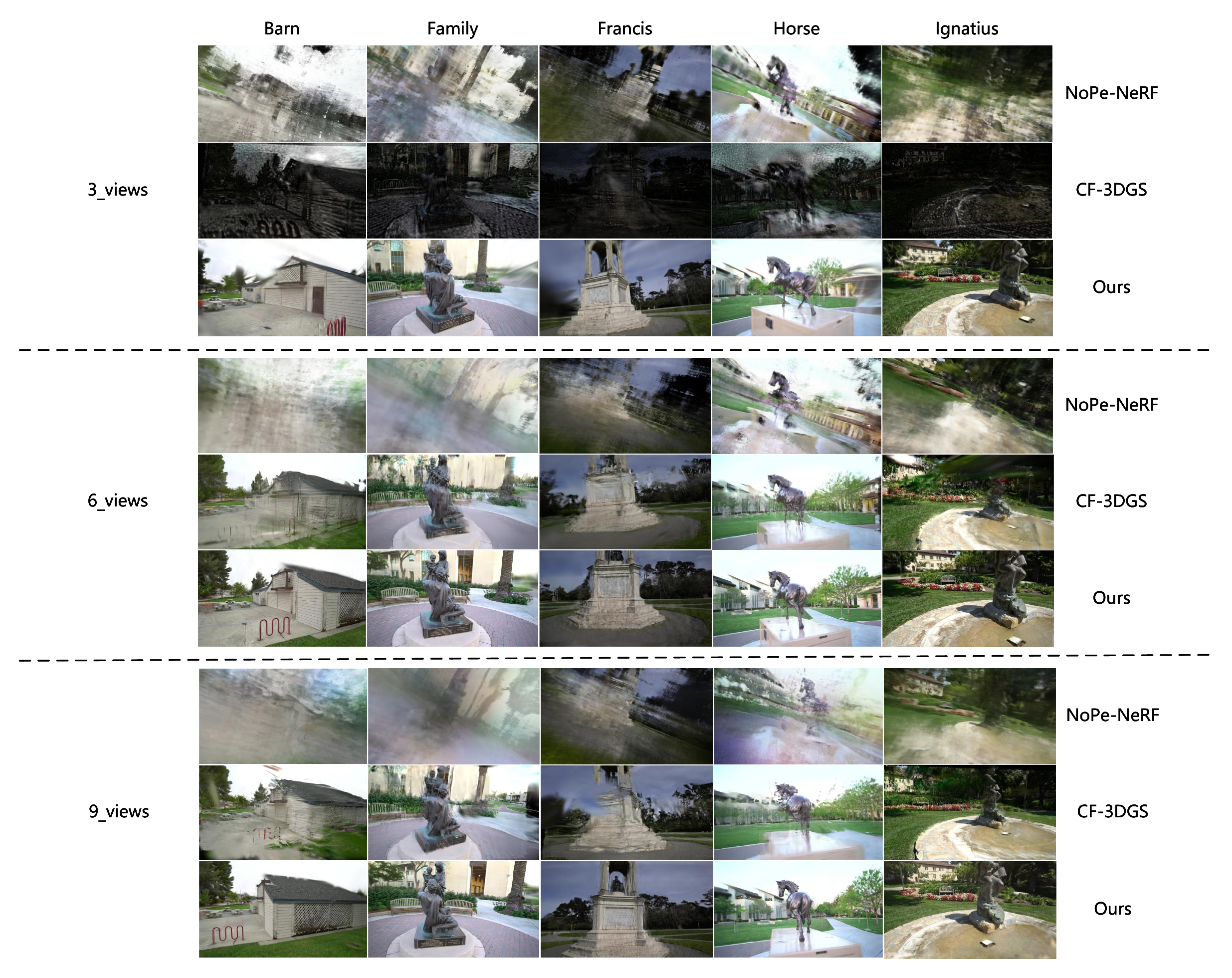}
\caption{ Qualitative comparison on TNT(3,6,9 views).The reconstruction of our method is more accurate and exhibits finer details.}
\label{fig:tnt}
\end{figure*}

To evaluate the performance of Gesplat on large-scale complex scenes, we conduct further comparisons on Tanks and Temples dataset. With a training setting of 6 views, Gesplat achieves the best results across all metrics as shown in Tab. \ref{tab:tnt_comparison}, demonstrating our robustness on challenging large-scale environments. We also visualize the qualitative comparisons in Fig. \ref{fig:tnt} to demonstrate the effectiveness of Gesplat.

\begin{table*}[htbp]
	\centering
	\caption{Quantitative comparison on Tank and Temples. All models are trained with six input views.The best result is highlighted in bold.}
	\setlength{\tabcolsep}{2mm}
	\footnotesize
    \resizebox{0.80\textwidth}{!}{
	\begin{tabular}{l|ccc|ccc|ccc}
		\toprule[1pt]
        \multirow{2}{*}{\textbf{scenes}} & \multicolumn{3}{c}{Ours} & \multicolumn{3}{c}{NoPe-NeRF\cite{bian2023nope}} & \multicolumn{3}{c}{CF-3DGS\cite{fu2024colmap}}  \\
        & PSNR$\uparrow$ & SSIM$\uparrow$ & LPIPS$\downarrow$  & PSNR$\uparrow$ & SSIM$\uparrow$ & LPIPS$\downarrow$ 
        & PSNR$\uparrow$ & SSIM$\uparrow$ & LPIPS$\downarrow$  \\
        \midrule
        Ballroom  &\textbf{20.641}&\textbf{0.662}&\textbf{0.231}&9.873&0.216&0.715&11.897&0.260&0.496 \\
        Barn      &\textbf{22.409}&\textbf{0.739}&\textbf{0.249}&12.885&0.507&0.726&14.235&0.491&0.493  \\
        Church    &\textbf{20.379}&\textbf{0.652}&\textbf{0.293}&14.301&0.355&0.680&13.629&0.350&0.576  \\
        Family    &\textbf{22.200}&\textbf{0.735}&\textbf{0.218}&11.961&0.429&0.775&14.617&0.514&0.497  \\
        Francis   &\textbf{23.487}&\textbf{0.695}&\textbf{0.324}&15.125&0.420&0.719&16.961&0.549&0.476  \\
        Horse     &\textbf{22.618}&\textbf{0.768}&\textbf{0.208}&10.629&0.397&0.705&16.797&0.617&0.383  \\ 
        Ignatius  &\textbf{22.563}&\textbf{0.669}&\textbf{0.264}&12.892&0.283&0.735&13.132&0.324&0.553  \\
        Museum    &\textbf{20.636}&\textbf{0.588}&\textbf{0.323}&11.872&0.285&0.764&15.798&0.495&0.439  \\
        \midrule
        mean      &\textbf{21.867}&\textbf{0.689}&\textbf{0.264}&12.442&0.362&0.727&14.633&0.450&0.489 \\
        \bottomrule[1pt]
        \end{tabular}}
	\label{tab:tnt_comparison}
	\vspace{-7pt}
\end{table*}

\subsection{Ablation Study}

We conduct a series of ablation studies to validate the effectiveness of each module in Gesplat and the rationality of parameter selection. Under a 3 views setting, we select the fortress scene from the LLFF dataset to establish our design choices. Quantitative and qualitative results of the ablation study are presented in Tab. \ref{tab:module_ablation} and Fig. \ref{fig:ablation}, respectively.

\begin{table}[htbp]
\caption{Ablations on the key modules. We evaluate them on the fortress scene of LLFF.} 
\centering
\setlength{\tabcolsep}{2mm}
	\footnotesize
\resizebox{0.4\textwidth}{!}{
\begin{tabular}{l|ccc}
    \toprule[1pt]
    \textbf{Method} & PSNR$\uparrow$ & SSIM$\uparrow$ & LPIPS$\downarrow$ \\
    \midrule
    Ours & \textbf{21.14} & \textbf{0.60} & \textbf{0.30} \\
    w/o Depth Reg. & 20.43 & 0.60 & 0.30 \\
    w/o Graph Opt. & 20.58 & 0.56 & 0.31 \\ 
    w/o Hybrid Rep. & 18.67 & 0.48 & 0.38 \\ 
    \bottomrule[1pt]
\end{tabular}}
\label{tab:module_ablation}
	\vspace{-7pt}
\end{table}

\subsubsection{Effect of hybrid representation (Hybrid Rep.)}

As shown in Tab. \ref{tab:module_ablation}, the removal of the hybrid representation leads to a drop in PSNR from 21.14 dB to 18.67 dB. As illustrated in Fig. \ref{fig:ablation}, the visual quality of the rendered image also degrades significantly, with noticeable holes and blurred texture of the table. These results demonstrate that, compared to the original Gaussian representation, introducing hybrid Gaussian representation based matching priors better preserves geometric information, improves structural completeness, and reduces the risk of overfitting in sparse view scenes.

\subsubsection{Effect of graph-guided optimization (Graph Opt.)}

Although graph-guided optimization is only applied in the last 200 iterations, its removal causes a quantitative decrease of 0.56 dB in PSNR, as shown in Tab. \ref{tab:module_ablation}. From the visual comparison in Fig. \ref{fig:ablation}, we observe that the edges of the floor become blurred and the details of the fortress are over-smoothed without graph-guided optimization. This indicates that graph-guided optimization effectively refines Gaussian attributes and enhances scene details.

\subsubsection{Effect of flow-based depth regularization (Depth Reg.)}

\begin{figure*}[htbp]
\centering 
\includegraphics[width=12cm,height=7cm]{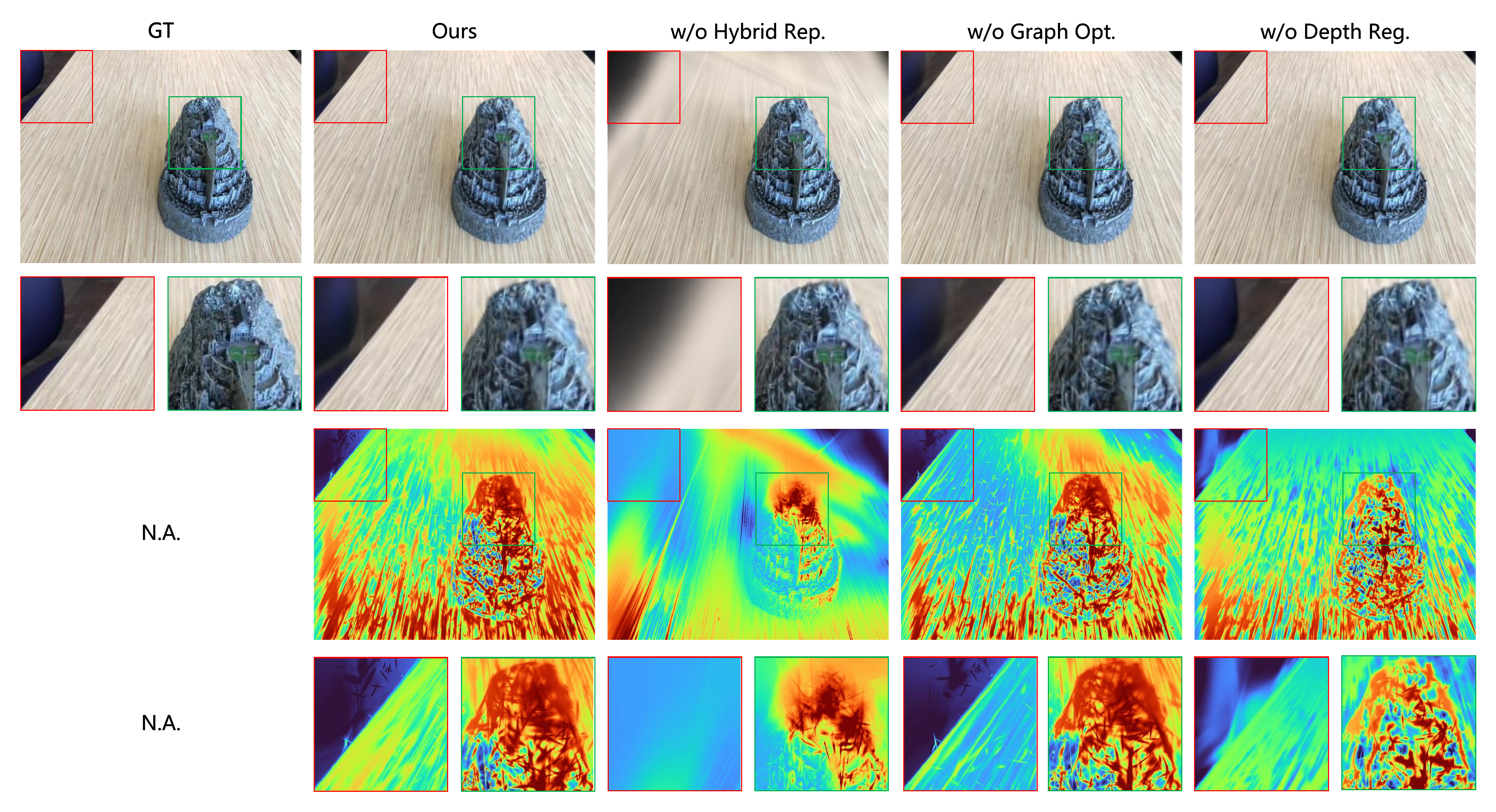}
\caption{ Ablation study on the key modules of our model.}
\label{fig:ablation}
\end{figure*}

Without flow-based depth regularization, it results in a slight decrease in PSNR, as reported in Tab. \ref{tab:module_ablation}. Compared to our full model, the absence of this module also blurs the edges of the fortress and fails to accurately reconstruct the gaps in the flooring, as shown in Fig. \ref{fig:ablation}. These results underscore the importance of this strategy.

\subsubsection{Effect of model parameter settings}

We further investigate the effectiveness of parameter selection using the trex scene from the LLFF dataset. The parameters under study include the weight of the depth loss $\lambda_{depth}$ and the regularization parameters for Gaussian attribute updates in the graph network: $\lambda_z$, $\lambda_s$, $\lambda_r$, $\lambda_c$, and $\lambda_\alpha$. The results are summarized in Tab. \ref{tab:parameter_setting}.The value of $\lambda_{depth}$ influences the geometric continuity of the scene. We find that $\lambda_{depth} = 0.1$ yields the best performance. Increasing $\lambda_{depth}$ leads to degradation across all metrics, indicating that excessive depth regularization over-smooths texture details and reduces reconstruction accuracy. Since we expect the graph network to refine scene details through Gaussian attribute updating, we place greater emphasis on SSIM and LPIPS during evaluation. Based on the results in Tab. \ref{tab:parameter_setting}, we set $\lambda_z = 0.1$, $\lambda_s = 0.1$, $\lambda_r = 0.05$, $\lambda_c = 0.01$, and $\lambda_\alpha = 1.0$.

\begin{table}[htbp]
\centering
\caption{Ablations on model parameter settings. We evaluate them on the trex scene of LLFF.}
\setlength{\tabcolsep}{2mm}
	\footnotesize
\resizebox{0.4\textwidth}{!}{
\begin{tabular}{cc|ccc}
    \toprule[1pt]
    \multicolumn{2}{c|}{\textbf{Parameter setting}} & PSNR$\uparrow$ & SSIM$\uparrow$ & LPIPS$\downarrow$\\
    \midrule
    \multirow{3}{*}{$\lambda_{depth}$} 
     & 0.1 & \textbf{19.532} & \textbf{0.649} & \textbf{0.318} \\
     & 0.2 & 19.102 & 0.623 & 0.360 \\
     & 0.3 & 17.173 & 0.570 & 0.392 \\
    \midrule
    \multirow{3}{*}{$\lambda_{z}$} 
     & 0.05 & \textbf{19.558} & 0.647 & 0.320 \\
     & 0.1 & 19.532 & \textbf{0.649} & \textbf{0.318} \\
     & 0.2 & 19.494 & 0.649 & 0.319 \\
    \midrule
    \multirow{3}{*}{$\lambda_{s}$} 
     & 0.01 & 19.505 & 0.647 & 0.321 \\
     & 0.1 & 19.532 & \textbf{0.649} & \textbf{0.318} \\
     & 1.0 & \textbf{19.576} & 0.647 & 0.318 \\
    \midrule
    \multirow{3}{*}{$\lambda_{r}$} 
     & 0.03 & \textbf{19.586} & 0.648 & 0.322 \\
     & 0.05 & 19.532 & \textbf{0.649} & \textbf{0.318} \\
     & 0.07 & 19.362 & 0.646 & 0.322 \\
    \midrule
    \multirow{3}{*}{$\lambda_{c}$} 
     & 0.01 & \textbf{19.532} & \textbf{0.649} & \textbf{0.318} \\
     & 0.1 & 19.451 & 0.643 & 0.320 \\
     & 1.0 & 18.679 & 0.641 & 0.416 \\
    \midrule
    \multirow{3}{*}{$\lambda_{\alpha}$} 
     & 1 & 19.532 & \textbf{0.649} & \textbf{0.318} \\
     & 2 & \textbf{19.596} & 0.647 & 0.320 \\
     & 3 & 19.498 & 0.647 & 0.322 \\
    \bottomrule[1pt]
\end{tabular}}
\label{tab:parameter_setting}
	\vspace{-7pt}
\end{table}

\section{Conclusion}

In this paper, we introduce Gesplat to address the challenge of sparse-view novel view synthesis from unposed images. Our framework first employs VGGT to generate initial dense point clouds and camera pose estimations. To enhance structural consistency, we introduce a hybrid Gaussian representation optimized with matching priors, enforcing geometric coherence through multi-view structural and rendering constraints. We further incorporate flow-based depth regularization to improve rendering accuracy and a graph-guided optimization module to refine Gaussian attributes for detailed scene recovery. Finally, we jointly optimize camera parameters and Gaussian representations. Extensive comparisons with other pose-free methods show that our approach achieves state-of-the-art performance on both forward-facing and large-scale scenes. A limitation remains when handling largely non-overlapping views, where matching priors become unreliable. Future work will explore more robust geometric constraints for extreme sparsity scenarios.

\bibliographystyle{elsarticle-num} 
\bibliography{refs}

\end{document}